\documentclass[conference]{IEEEtran}
\IEEEoverridecommandlockouts
\usepackage{graphicx} 
\usepackage{pifont}
\usepackage{amsmath}
\usepackage{amssymb}  
\usepackage{amsfonts}
\usepackage{algorithmic}
\usepackage{array}
\usepackage{multicol}

\usepackage{textcomp}
\usepackage{stfloats}
\usepackage{url}
\usepackage{verbatim}
\usepackage{graphicx}
\usepackage{caption}
\usepackage{subcaption}
\usepackage[ruled, vlined, linesnumbered]{algorithm2e}
\usepackage{cite}
\usepackage{booktabs}
\usepackage{amssymb,amsfonts,bm}
\usepackage{amsmath,tikz}
\usepackage{color}
\usepackage{mathtools}
\usepackage[hidelinks]{hyperref} 
\usepackage{rotating}
\usepackage{blkarray}
\usepackage{physics}
\usetikzlibrary{arrows}
\usepackage{makecell}

\usepackage{amsthm}
\usepackage{comment}
\usepackage{multirow}
\usepackage{geometry}
\graphicspath{{images/}}

\begin{document}

\title{GERA: Geometric Embedding for Efficient Point Registration Analysis
}


\author{
Geng Li\textsuperscript{1,2}, 
Haozhi Cao\textsuperscript{1}, 
Mingyang Liu\textsuperscript{2},
Shenghai Yuan\textsuperscript{1},
Jianfei Yang\textsuperscript{1,*}
\thanks{\textsuperscript{1}Nanyang Technological University, Singapore.}%
\thanks{\textsuperscript{2}Shandong University, China.}
\thanks{\textsuperscript{*}Corresponding author: {\tt\small jianfei.yang@ntu.edu.sg}}
}

\maketitle

\begin{abstract}
Point cloud registration aims to provide estimated transformations to align point clouds, which plays a crucial role in pose estimation of various navigation systems, such as surgical guidance systems and autonomous vehicles. Despite the impressive performance of recent models on benchmark datasets, many rely on complex modules like KPConv and Transformers, which impose significant computational and memory demands. These requirements hinder their practical application, particularly in resource-constrained environments such as mobile robotics. In this paper, we propose a novel point cloud registration network that leverages a pure MLP architecture, constructing geometric information offline. This approach eliminates the computational and memory burdens associated with traditional complex feature extractors and significantly reduces inference time and resource consumption. Our method is the first to replace 3D coordinate inputs with offline-constructed geometric encoding, improving generalization and stability, as demonstrated by Maximum Mean Discrepancy (MMD) comparisons. This efficient and accurate geometric representation marks a significant advancement in point cloud analysis, particularly for applications requiring fast and reliability.
\end{abstract}


\section{Introduction}
Point cloud registration is a fundamental problem in 3D scene understanding and robotics. The goal is to acquire the point-wise transformation matrix that aligns two point clouds, which plays a vital role in various downstream tasks (e.g., motion estimation~\cite{liu2019flownet3d,shen2023self}, 3D scene reconstruction~\cite{chang2024gaussreg}, SLAM~\cite{cattaneo2022lcdnet,lim2024quatro++}, and surgical navigation~\cite{baum2021real}).

\begin{figure}[htbp]
    \centering
    \includegraphics[width=.9\linewidth]{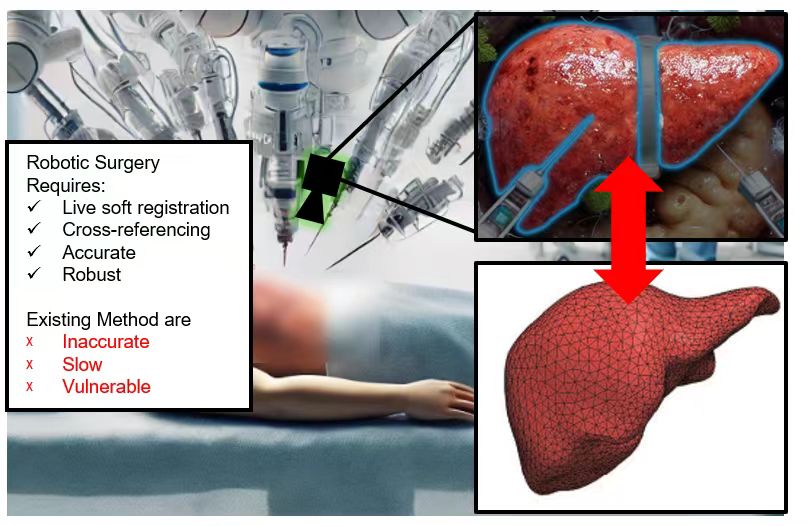}
    \caption{In robot-assisted surgery, real-time analysis and registration of target organs are required, which presents new challenges for robustness and efficiency.}
    \label{fig:intro}
\vspace{-10pt}
\end{figure}

Recent non-learning-based methods for non-rigid registration perform well but are hindered by high inference time, limiting real-time use. Learning-based methods, which eliminate complex optimization at test time, improve efficiency but still fall short of real-time performance, necessitating further enhancements for practical deployment~\cite{li2022lepard,qin2022geometric}. A typical example is the point cloud registration of organs in surgical robots~\cite{taylor2022surgical,ringel2024comparing} as shown in Fig.~\ref{fig:intro}. On the one hand, the registration algorithm, in this case, needs to be robust and highly accurate to guarantee the success of the surgery. On the other hand, it must satisfy the real-time inference requirement even with rather limited computational resources (e.g., embedded devices). Therefore, developing accurate and efficient learning-based point cloud registration methods is significant, bridging the gap between benchmarking and real-life deployment. 



Existing learning-based solutions~\cite{qin2022geometric,li2022lepard} rely on complex local geometric feature extractors that extract usable registration features from raw 3D points~\cite{thomas2019kpconv,vaswani2017attention}. Despite their effectiveness, these powerful extractors (e.g., KPConv~\cite{thomas2019kpconv} and Transformer~\cite{vaswani2017attention}) usually contain masses of trainable parameters, bringing significant computational expenses during both the training and inference stages. As the model becomes increasingly complex, its performance improvements fall short of meeting the demands for real-time processing.

 We revisited recent learning-based registration methods and found that their primary efficiency bottleneck stems from feature extraction and online local construction modules~\cite{qin2022geometric,qi2017pointnet++,li2022lepard}. Drawing inspiration from non-learning methods~\cite{serafin2015nicp}, we propose replacing traditional positional information with geometric features,  to optimize registration performance. Our approach constructs information-rich geometric-encoded inputs, which offer two key advantages. First, by using efficient encoding algorithms, geometric 3D representations can be extracted without learnable modules, reducing the computational load during both offline training and online inference. Second, these geometric representations are more informative than raw 3D points, leading to better performance even with simpler learnable modules.

Based on the above analysis, in this paper, we propose GERA, a method for \underline{GE}ometric embedding for lea\underline{R}ning-based efficient point registr\underline{A}tion with a lightweight trainable network, improving both the effectiveness and efficiency of registration. Given raw point clouds as input, we first construct a 3D descriptor for each point by forming a fully connected graph using this point and its neighboring points. The edges of the graph represent the distances between every pair of points.
We conducted a kernel-based statistical analysis using Maximum Mean Discrepancy(MMD)~\cite{dziugaite2015training} to demonstrate the stability and robustness of our geometric embedding encoding.
which is a kernel-based statistical measure employed to analyze the similarity of encoded embeddings between different input samples. By leveraging the MMD, we demonstrated that the stability of geometric information encoding features is superior to previous state-of-the-art learning-based solutions. Additionally, our descriptors can be constructed offline, significantly reducing training and inference time. Following prior work that employs MLP architectures for point cloud analysis and registration, we utilize this simple network architecture to effectively leverage the informative descriptors, thereby significantly surpassing previous complex learning-based solutions. Our experiment results show that the inference speed was increased by 22x, resulting in a 115\% improvement in prediction accuracy compared to the existing state-of-the-art solution.




Our contributions can be summarized as follows:
\begin{itemize}
    \item We propose an offline method that efficiently and accurately constructs geometric information from point clouds.
    \item We are the first to replace 3D coordinate input with geometric encoding information in point cloud processing.
   \item We surpassed the SOTA by 12.5\% while using only 3\% of the computation time required by existing methods.
\end{itemize}

\section{Related work}

\subsection{Feature Encoding}
Feature encoding for point cloud data began with PointNet~\cite{qi2017pointnet}, which utilized rotation-invariance and an MLP architecture but lacked the ability to capture local geometric information. This was improved by PointNet++~\cite{qi2017pointnet++}, which introduced hierarchical feature extraction from local to global levels, and DGCNN~\cite{wang2019dynamic}, which considered the relative distances between neighboring points. KPConv~\cite{thomas2019kpconv} further advanced this by extracting features within a spherical range, similar to convolution. In point cloud registration, methods like GeoTransformer~\cite{qin2022geometric} and Lepard~\cite{li2022lepard} adopted transformer architectures to effectively model local geometric information, though with increased computational overhead.

\subsection{Non-learning-based Methods}
The Iterative Closest Point (ICP)~\cite{besl1992method} algorithm based on the least squares method relies heavily on initialization and converges slowly. Coherent Point Drift~\cite{myronenko2010point} (CPD) formulates the registration task as a probability density estimation problem, iteratively updating point correspondences and non-rigid transformation parameters using the expectation-maximization technique. However, CPD requires multiple iterations and is sensitive to occlusions and outliers. Methods employing Graph-Laplacian regularization~\cite{panaganti2015robust} and context-aware Gaussian fields (SCGF)~\cite{wang2017robust} improve correspondence and transformation estimation but depend on high-quality assignment initialization. Bayesian Coherent Point Drift (BCPD)~\cite{hirose2020bayesian} enhances CPD's convergence through variational Bayesian inference but remains prone to local minima. The MR-RPM~\cite{ma2017non} method based on manifold regularization captures the intrinsic geometry of point sets using manifold regularization priors. PointSetReg~\cite{zhao2024correspondence} utilizes unsupervised clustering analysis, avoiding explicit point correspondences to improve robustness and efficiency. However, the heavy reliance of PointSetReg on cluster-level information may reduce precision when handling complex point clouds.

\subsection{Learning-based Methods}
FPT~\cite{baum2021real} utilizes a decoder composed of PointNet and fully connected layers to perform non-rigid registration on prostate point clouds, offering high efficiency but limited robustness due to the lack of a geometric construction module. Lepard~\cite{li2022lepard} employs a transformer architecture to form attention heads between point clouds, extracting the point correspondence matrix, which is then input into algorithms such as N-ICP~\cite{serafin2015nicp} for non-rigid registration. While Lepard provides comprehensive encoding of point cloud information, its slower processing speed limits practical applications. Similar to non-rigid registration, the problem of scene flow estimation has garnered increasing attention in recent years. PointPWC-Net~\cite{wu2020pointpwc} refines the flow iteratively at multiple scales, enhancing its ability to capture fine-grained motion details, but its computational complexity remains relatively high. BI-PointFlowNet~\cite{cheng2022bi} incorporates bidirectional learning for point clouds, enhancing accuracy and robustness, but it encounters difficulties when handling scenes with large displacements. DiffFlowNet~\cite{liu2024difflow3d} utilizes a powerful and complex diffusion model, achieving millimeter-level error in scene flow estimation, but it also introduces the challenge of prolonged inference time.

\section{METHODOLOGY}
\noindent
\textbf{Probelm Definition:}
In non-rigid registration, we are given a source point set \( \mathbf{P}_\mathcal{S} = \{x_1, \dots, x_i, \dots, x_M\} \) and a target point set \( \mathbf{P}_\mathcal{T} = \{y_1, \dots, y_j, \dots, y_N\} \), where \( x_i, y_j \in \mathbb{R}^3 \) represent the 3D coordinates of the points, and \( M, N \) are their respective counts. The goal is to estimate displacement vectors \( \mathbf{D}_{\text{pred}} = [\mathbf{d}_{\text{pred}_1}, \dots, \mathbf{d}_{\text{pred}_M}] \), which deform each point \( x_i \in \mathbf{P}_\mathcal{S} \) to align with \( \mathbf{P}_\mathcal{T} \).

The deformed point set \( \mathbf{P}_\mathcal{S}' = \{x'_1, \dots, x'_i, \dots, x'_M\} \) is computed as:
\begin{equation}
x'_i = x_i + f(x_i, \mathbf{D}_{\text{pred}}) = x_i + \mathbf{d}_{\text{pred}_i} + \epsilon(x_i),
\end{equation}
where \( x'_i \) is the deformed point in \( \mathbf{P}_\mathcal{S}' \), and \( \epsilon(x_i) \) represents a small adjustment term to enhance registration smoothness.

We propose GERA, which enhances point cloud registration by leveraging offline 3D geometric representations. By embedding neighbor information in a point-wise manner, our method achieves more stable encoding than 3D coordinates, as shown by MMD~\cite{dziugaite2015training} analysis. Therefore, GERA can significantly surpass existing solutions with more lightweight learnable modules, as shown in Fig.~\ref{fig:ovearll_method}

\subsection{Efficient Offline Geometric Representations }
Previous studies on local geometric information used 3D spatial coordinates of point clouds with online extraction, resulting in a computational complexity of $O(n)$, where $n$ is the number of training epochs, demanding significant computational and memory resources. This paper instead proposes an offline geometric information construction method with $O(1)$ complexity, significantly alleviating the overhead during the training process. Based on our quantitative analysis with MMD, we further demonstrate that our proposed geometric representations possess superior generalizability and stability compared to others.

\begin{figure}[htbp]
    \centering
    \includegraphics[width=0.8\linewidth]{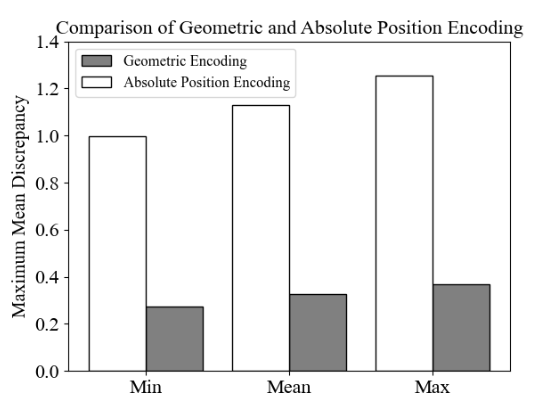}
    \caption{Perform MMD analysis on the liver dataset using a batch size of 32 for both absolute position encoding and geometric encoding}
    \label{fig:mmd_comp}
\end{figure}

\subsubsection{Construction of Geometric Representations}
Our method improves upon previous local geometric information construction by using a simple geometric prior. Traditional methods~\cite{qin2022geometric} compute distances and angles between points and neighbors, which are not directly comparable to MLP input because angles and distances represent different dimensions of information. By forming a triangle with the original point and its two neighbors, and defining it by side lengths, we reduce the geometric relationships to a consistent and usable format for the network, which can be formulated as:
\begin{equation}
\begin{aligned}
\alpha &= \| \mathbf{P}_i - \mathbf{P}_j \|, & \beta &= \| \mathbf{P}_i - \mathbf{P}_k \|, \\
\gamma &= \| \mathbf{P}_j - \mathbf{P}_k \|, & \mathbf{G}_{(i,j,k)} &= \text{concat}(\alpha, \beta, \gamma),
\end{aligned}
\end{equation}
where \(\mathbf{P}_i\), \(\mathbf{P}_j\), \(\mathbf{P}_k\) are the coordinates of points, and \(\mathbf{G}_{(i,j,k)}\) donates the geometric information constructed for points \(i,j,k\). $\text{concat}(\cdot)$ indicates the concatenation operation.

Using only two neighboring points for geometric construction forms the simplest fully connected graph, yielding limited information. To further enrich the embedded information, we consider each point's \(n\) neighbors as vertices of a fully connected graph, resulting in \(C_n^2\) pairwise distances. This comprehensive modeling retains only the edge lengths, which, according to distance geometry and rigidity theory, uniquely determine a 3D structure~\cite{mucherino2012distance}. This approach is widely used in structural chemistry and drug design~\cite{cassioli2015algorithm,lavor2019minimal}, which could be formulated as:
\begin{equation}
 \mathbf{sc}^{i} \in \mathbb{R}^d = \text{concat}(\{ \mathbf{G}_{(i,j,k)}\mid 1 \leq j < k \leq n\}),
\end{equation}
where $\mathbf{sc}^{i}$ denotes the fully connected graph constructed from $n$ neighboring points. $d$ represents the total number of edges in the fully connected graph, \(C_n^2\).

\subsubsection{Analysis of Stability and Generalization}
Non-rigid point cloud registration entails more complex transformations than rigid registration, resulting in flexible  data distribution variations. Thus, a robust feature representation with strong generalization is essential. To assess the stability and generalization of geometric encoding, we performed a quantitative analysis based on MMD, a criterion widely used to measure discrepancy in domain adaptation~\cite{cao2024mopa,kong2023conda,rochan2022unsupervised}.

Specifically, MMD measures the distance between samples in a high-dimensional feature space, with smaller distances indicating greater similarity and thus stronger generalization and stability. In this context, MMD can provide a quantitative assessment about the similarity of the encoded features between samples, expressed as:
\begin{equation}
\text{MMD}^2( \mathbf{P}_\mathcal{S}, \mathbf{P}_\mathcal{T}) = \left\| \mathbb{E}_{x \sim \mathbf{P}_\mathcal{S}}[\phi(x)] - \mathbb{E}_{y \sim \mathbf{P}_\mathcal{T}}[\phi(y)] \right\|_{\mathcal{H}}^2,
\end{equation}
where \(\phi(x)\) is the mapping function that projects the \(x\) into a high-dimensional Reproducing Kernel Hilbert Space \(\mathcal{H}\).

\begin{figure*}[t]
\centering
\includegraphics[width=.9\textwidth]{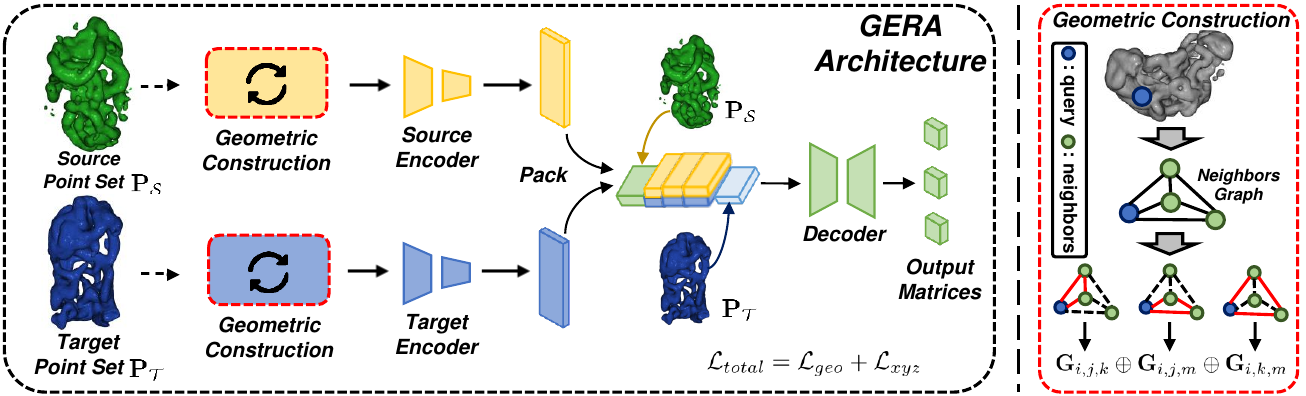}
\caption{The ovearll structure of GERA. The raw point cloud coordinate information is processed by an offline geometric constructor, resulting in encoded geometric information. This encoded data is passed through an MLP-based encoder, extracting $\mathcal{F}^{\mathit{geo}}_{\mathbf{P}_\mathcal{S}}$ and $\mathcal{F}^{\mathit{geo}}_{\mathbf{P}_\mathcal{T}}$. Subsequently, the original coordinates are concatenated with the geometric information, yielding a fully populated feature map \(\mathcal{F}_\mathbf{full}^{\mathit{geo}} \in \mathbb{R}^{d \times n}\). Finally, the \(\mathcal{F}_\mathbf{full}^{\mathit{geo}} \) is fed into a decoder, to obtain the displacement matrix.}
\label{fig:ovearll_method}
\vspace{-10pt}
\end{figure*}

By expanding and simplifying equation (3), we can derive the computation formula for MMD:
\begin{equation}
\begin{aligned}
\text{MMD}^2(\mathbf{P}_\mathcal{S}, \mathbf{P}_\mathcal{T}) &= \mathbb{E}_{x,x' \sim \mathbf{P}_\mathcal{S}}[k(x,x')] + \mathbb{E}_{y,y' \sim \mathbf{P}_\mathcal{T}}[k(y,y')] \\
&\quad - 2\mathbb{E}_{x \sim P, y \sim Q}[k(x,y)],
\end{aligned}
\end{equation}
\begin{equation}
k(x, y) = \exp\left(-\frac{\|x - y\|^2}{2\sigma^2}\right),
\end{equation}
where \(x\) and \(x'\) represent two independent samples drawn from the distribution \(\mathbf{P}_\mathcal{S}\), and \(k(\cdot, \cdot)\) is the kernel function. Equation (5) specifies the Gaussian kernel function.

In this study, we employed identical encoder architectures for both 3D coordinate encoding and geometric encoding as inputs to MMD. As shown in Fig.~\ref{fig:mmd_comp}, the MMD values produced by geometric encoding were significantly lower than those from position encoding, indicating that GERA’s geometric encoding method is more stable and has stronger generalization capabilities. Moreover, the minimum, mean, and maximum MMD values for geometric encoding were 0.2736, 0.3178, and 0.3668, respectively, which are notably lower than the corresponding values of 0.9968, 1.1702, and 1.2527 for position encoding. This demonstrates that our method can achieve robust and reliable feature extraction across different samples.

\subsection{GERA architecture}

\subsubsection{Geometric Encoder and Decoder}
As we discussed, offline geometric encoding extrated rich features from the raw point clouds. By leveraging these features, we can effectively and efficiently address the non-rigid registration task with a simple MLP architecture. Given the raw point cloud inputs \(\mathbf{P}_\mathcal{S}\) and \(\mathbf{P}_\mathcal{T}\), the geometric extraction process \(\mathit{Geo(\cdot)}\) first constructs the geometric information. The resulting geometric embeddings from \(\mathbf{P}_\mathcal{S}\) and \(\mathbf{P}_\mathcal{T}\) can be expressed as:
\begin{equation}
\begin{aligned}
, \\
\mathbf{P}_\mathcal{T}^{\mathit{geo}} &= \mathit{Geo}(\mathbf{P}_\mathcal{T}) = \left\{ \mathbf{sc}_{\mathbf{P}_\mathcal{T}}^{i} \in \mathbb{R}^d \mid 1 \leq i \leq m \right\},
\end{aligned}
\end{equation}
where \(\mathbf{sc}\) denotes the edge length in the encoded fully connected graph, \(i,k\) are the point indices, and \(m\) represents the number of points in the point cloud.

The feature extraction of $\mathbf{P}_\mathcal{S}$ and $\mathbf{P}_\mathcal{T}$ is performed through an MLP, resulting in geometric features $\mathcal{F}^{\mathit{geo}}_{\mathbf{P}_\mathcal{S}}$ and $\mathcal{F}^{\mathit{geo}}_{\mathbf{P}_\mathcal{T}}$, respectively. These two features are then concatenated to generate a new representation $\text{concat}(\mathcal{F}^{\mathit{geo}}_{\mathbf{P}_\mathcal{S}},\mathcal{F}^{\mathit{geo}}_{\mathbf{P}_\mathcal{T}})$. This process can be formulated as:
\begin{equation}
\mathcal{F}^{\mathit{geo}} = \text{concat}\left\{ \mathbf{P}_\mathcal{S}, \lambda(\text{concat}(\mathcal{F}^{\mathit{geo}}_{\mathbf{P}_\mathcal{S}},\mathcal{F}^{\mathit{geo}}_{\mathbf{P}_\mathcal{T}}), n), \mathbf{P}_\mathcal{T} \right\},
\end{equation}
where \(\lambda(\rho, n)\) represents the replication of \(\rho\) \(n\) times.

The concatenated feature vector \(\mathcal{F}^{\mathit{geo}} \in \mathbb{R}^d\) is replicated \(n\) times, where \(n\) is the number of points in the point cloud, producing a fully populated feature map \(\mathcal{F}_\mathbf{full}^{\mathit{geo}} \in \mathbb{R}^{d \times n}\). This feature map concatenates in the packing stage with the point cloud coordinates. The decoder, an MLP, then processes this combined data to output the predicted displacement matrix for point cloud registration.

\begin{table*}[ht]
\centering
\renewcommand{\arraystretch}{1.1} 
\setlength{\tabcolsep}{8pt} 
\caption{\small Comparison of Various Methods on Liver, Brain and Stomach Point Cloud Datasets. The best results are in \textbf{bold} while the second best are \underline{underlined}.}
\begin{tabular}[width=\textwidth]{l c c c c c c c c}
\hline\hline
\textbf{Method} & \textbf{Training} & \textbf{Years} & \textbf{RMSE (mm)} & \textbf{CD (mm)} & \textbf{TT (s)}& \textbf{IT (ms)} & \textbf{FLOPs (G)} & \textbf{Parameters (M)} \\
\hline
BCPD~\cite{hirose2020bayesian} & - & 2020 & 17.75 & 11.17 &-& 5105.63 & - & - \\
PointSetReg~\cite{zhao2024correspondence} & - & 2024 & \textbf{7.07} & \textbf{2.71} & - &718.46& - & - \\
\hline
FPT\cite{baum2021real} & Sup. & 2021 & 12.80 & 8.08 &30.48& 8.23 & 7.58 & 5.61 \\
PointPWC\cite{wu2020pointpwc} & Sup. & 2020 & 10.01 & 6.41 &71.97& 26.12 & 8.91 & 7.72 \\
DifFlow3D\cite{liu2024difflow3d} & Sup. & 2024 & 13.51 & 8.15 &155.26& 53.39 & 16.97 & 3.51 \\
Bi-pointflownet\cite{cheng2022bi} & Sup. & 2022 & 8.07 & 5.51 &72.06& 29.87 & 8.16 & 7.96 \\
Lepard\cite{li2022lepard} & Sup. & 2022 & 8.10 & 6.02 & 364.98&109.38 & 40.62 & 23.72 \\
\hline
\textbf{GERA-xyz} (Ours) & Sup. & - & 8.13 & 7.88 &\textbf{26.10}& \textbf{6.16} & \textbf{3.19} & \textbf{2.96} \\
\textbf{GERA-geo} (Ours) & Semi. & - & \underline{7.01}& \underline{5.48} &\underline{26.34}& \textbf{6.16} & \textbf{3.19} & \textbf{2.96} \\
\hline\hline
\end{tabular}
\label{tab:liver_results}
\end{table*}

\subsubsection{Loss function}
Previous methods used 3D coordinates as a metric to quantify the differences between two point clouds. However, we found that the geometric information constructed by GERA can capture detailed geometric differences in the point clouds. Therefore, we introduced \(\mathcal{L}_{geo}\), an unsupervised loss function that can be combined with existing registration methods, leading to significant improvements.

Our \(\mathcal{L}_{geo}\) is formed by extracting geometric information from the target and output point clouds using the geometric extraction
process \( \mathit{Geo(\cdot)} \), with the error calculated via an adapted Chamfer distance loss.
\begin{equation}
\begin{split}
   \mathcal{L}_{geo} = & \sum_{\mathbf{p} \in \mathbf{P}_\mathcal{S}} \min_{\mathbf{q} \in \mathbf{P}_\mathcal{T}} \|\mathit{Geo}\mathbf{(p)} - \mathit{Geo}\mathbf{(q)}\|^2 \\
   + & \sum_{\mathbf{q} \in \mathbf{P}_\mathcal{T}} \min_{\mathbf{p} \in \mathbf{P}_\mathcal{S}} \|\mathit{Geo}\mathbf{(q)} - \mathit{Geo}\mathbf{(p)}\|^2,
\end{split}
\end{equation}
where \(\mathbf{p}\) and \(\mathbf{q}\) denote points in the source point cloud and the target point cloud.

Additionally, we employed the commonly used root mean square error (RMSE) to calculate the error between the source and target point clouds, defining it as \(\mathcal{L}_{xyz}\). During training, \(\mathcal{L}_{geo}\) and \(\mathcal{L}_{xyz}\) form a semi-supervised combination learning framework, and the total loss can be expressed as:

\begin{equation}
    \mathcal{L}_{total} = \alpha\mathcal{L}_{geo} + (1-\alpha)\mathcal{L}_{xyz},
\end{equation}
where \(\alpha = 0\) corresponds to our proposed GERA-xyz, while \(\alpha \neq 0\) corresponds to GERA-geo.

\begin{figure*}[t]
\centering
\includegraphics[width=0.95\textwidth]{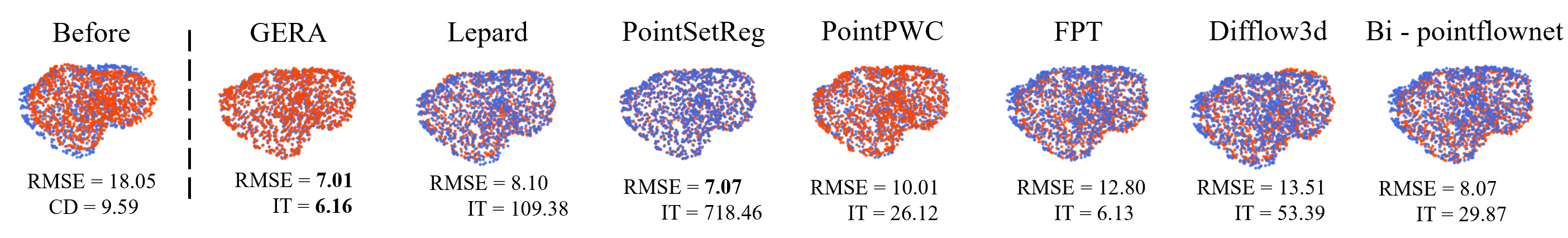}
\caption{The qualitative registration results on the dataset where the two point sets exhibit pure deformation. The blue and red point sets represent the source and target point clouds, respectively. For both Case 1 and Case 2, the maximum noise magnitude is 2 mm, and the deformation magnitude is 18 mm.}
\label{fig:liver}
\end{figure*}

\section{ EXPERIMENTS AND RESULTS}
\subsection{Dataset}
To simulate realistic surgical scenarios, we selected the point cloud subsets of the liver, brain, stomach, and small bowel from the MedShapeNet dataset~\cite{li2023medshapenet}. The original point clouds, consisting of 10,000 points, were downsampled to 1,024 points to mimic the sparse point clouds commonly encountered in real-world scenarios. Two benchmarks are constructed, where the first contains 1,033 point clouds of livers, brains, and stomachs, while the other consists of 131 small bowel samples. The latter poses a significant challenge because of its large variability and the limited number of samples. For both benchmarks, the samples were divided into training, validation, and test sets with a ratio of 8: 1: 1. The intraoperative source point sets were generated using the Thin Plate Spline (TPS)~\cite{bookstein1990thin} method, with a deformation value set at 19 mm to meet the requirements for organ registration. To simulate realistic scenarios, random noise was introduced ranging from 1 to 3 mm.

\subsection{Evaluation metrics}
To evaluate the registration quality, we utilize two
different evaluation metrics, i.e., RMSE and Chamfer distance(CD)~\cite{nguyen2021point,baum2021real}.
In addition to quality, we also evaluate the efficiency through various metrics. Training time (TT) represents the time required for each model to train a single epoch. In practical applications, when encountering unseen classes, retraining may be necessary. Therefore, shorter training times are advantageous for the real-world deployment of models. Inference time(IT) refers to the time required by a trained model to process a single-point cloud during testing. To achieve real-time performance, the inference time should be minimized as much as possible. In addition, FLOPs and the parameter numbers are evaluated to measure the complexity and computational efficiency of the model.

\subsection{Implementation Details}

All methods were implemented using the PyTorch framework on a single GPU (Nvidia GeForce RTX 4090, 24GB). The model was further fine-tuned on 1024 points randomly sampled from the original point sets, each consisting of 10,000 points. The Adam optimizer was employed, with a batch size set to 1, and the network was trained for a total of 300 epochs. For the comparative methods, we utilized their publicly available code versions and setup for epochs, optimizers, and hyperparameters.

To ensure fairness, all comparative methods have been thoroughly retrained on our organ datasets.

\subsection{ Experiment Results}
\subsubsection{ Comparisons with Other Methods on The Combined Dataset}
Fig. ~\ref{fig:liver} compares our method with other approaches. While FPT~\cite{baum2021real}, PointPwc~\cite{wu2020pointpwc}, Difflow3D~\cite{liu2024difflow3d}, and Bi-pointflownet~\cite{cheng2022bi} have faster inference times, their registration errors of 12.80 mm, 10.01 mm, 13.51 mm, and 8.07 mm do not meet accuracy or real-time requirements. PointSetReg~\cite{zhao2024correspondence} achieves a lower error but processes only one frame per second, limiting its efficiency. In contrast, our method achieves the best result among learning-based methods with an error of 7.01 mm and real-time performance at 156 frames per second.

Table~\ref{tab:liver_results} shows quantitative results on the combined dataset of liver, brain, and stomach using RMSE. FPT yields a 13.4 mm error, while PointPWCNet~\cite{wu2020pointpwc} and Bi-pointflownet~\cite{cheng2022bi} achieve 10.01 mm and 8.07 mm, but their inference times remain too slow for real-time use. Lepard~\cite{li2022lepard} reduces the error to 8.10 mm but also fails to meet real-time demands. PointSetReg~\cite{zhao2024correspondence} achieves a 7.07 mm error but requires 718.46 ms per point cloud. Our method, with a 7.01 mm error, performs real-time registration at 250 frames per second with the lowest FLOPs, at 20 percent of other methods. It also fine-tunes quickly, completing retraining in a fraction of the time needed by Bi-pointflownet~\cite{cheng2022bi} and Lepard~\cite{li2022lepard}, demonstrating superior efficiency and adaptability.

\begin{figure*}[t]
\centering
\includegraphics[width=0.98\textwidth]{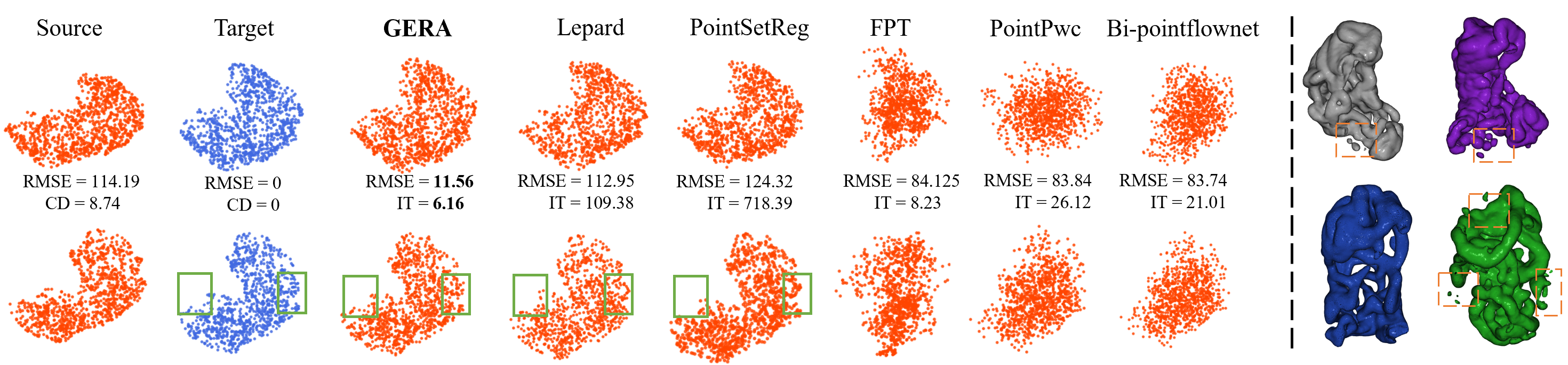}
\caption{The qualitative registration results on the dataset show that the two point sets exhibit pure deformation. The red and blue point sets represent the source and target point clouds, respectively. The deformation magnitude is 114.19 mm. On the right side of the figure, four randomly selected samples from the small intestine dataset are displayed, with yellow boxes marking breakpoints and noise locations.}
\label{fig:small_bowel}
\end{figure*}

\subsubsection{ Results on Small Bowel Dataset}
In robotic surgery, registration tests on multiple organs are often required. Additionally, point clouds collected from medical organs frequently contain substantial noise and are often incomplete in shape. To evaluate the robustness and capability of GERA, we conducted experiments on a more realistic and challenging small intestine dataset. These challenges arise from two main factors: first, the structure and distribution of the small intestine are more complex and diverse; second, the dataset is limited, containing only 131 samples, which accounts for 13\% of the entire dataset. Fig. 5 qualitatively presents four small bowel samples with significant distribution differences, with noise and breakpoints highlighted in yellow boxes.
\begin{table}[h]
\centering
\renewcommand{\arraystretch}{1.1} 
\setlength{\tabcolsep}{5pt} 
\caption{\small Benchmark Comparison of Various Methods. The best results are in \textbf{bold}, while the second best are \underline{underlined}.}
\begin{tabular}{l c c c c}
\hline\hline
\textbf{Method} & \textbf{RMSE (mm)} & \textbf{CD (mm)} & \textbf{IT (ms)} & \textbf{TT(s)} \\
\hline
PointSetReg &124.32 & \textbf{4.52} & 718.39& - \\
\hline
Bi-pointflownet\cite{cheng2022bi} &83.74 & 21.01 & 29.87 & 9.32 \\
PointPWC\cite{wu2020pointpwc} & 83.84& 20.92 & 26.12 & 9.14 \\

Lepard\cite{li2022lepard}  & 112.95 & 9.03 & 109.38 & 193.12 \\
FPT\cite{baum2021real} & 84.13 & 40.49 & 8.23 & 16.13 \\
\hline
\textbf{GERA-xyz} & \underline{12.87} & 7.94 & \textbf{6.16} & \textbf{3.02} \\
\textbf{GERA-geo} & \textbf{11.56} & \underline{7.74} & \textbf{6.16} & \underline{3.09} \\
\hline\hline
\end{tabular}
\label{tab:results_ii}
\end{table}
Fig. ~\ref{fig:small_bowel} compares the performance of our method with other approaches. Scene flow methods such as FPT~\cite{baum2021real}, PointPwc~\cite{wu2020pointpwc}, and Bi-pointflownet~\cite{cheng2022bi} fail to perform effective point cloud registration, resulting in transformed point clouds that appear as scattered points. Although Lepard~\cite{li2022lepard} and PointSetReg~\cite{zhao2024correspondence} maintain the shape of the small bowel point clouds, both methods tend to cluster points, leading to significantly higher RMSE values. In contrast, our method achieves the best registration results by aligning the transformed point cloud with the target point cloud while requiring the shortest inference time.

Table~\ref{tab:results_ii} presents the quantitative results on the small bowel dataset using RMSE. We tested only those methods that performed well on the liver dataset. The FPT~\cite{baum2021real} method has an error of 84.13 mm, while PointPWCNet~\cite{wu2020pointpwc} and Bi-pointflownet~\cite{cheng2022bi} achieve errors of 83.84 mm and 83.74 mm, respectively. However, all three methods perform disastrously on the small intestine dataset and fail to complete the registration task. Although Lepard~\cite{li2022lepard} and PointSetReg~\cite{zhao2024correspondence} show RMSE of 112.95 mm and 124.32 mm, respectively, due to their tendency to cluster points together, they maintain the basic shape of the point cloud and perform well in the CD metric, with values of 9.03 mm and 4.52 mm. Nonetheless, their inference times remain excessively long, severely impeding real-time point cloud registration. In contrast, our method achieves an error of 11.56 mm and is the only method capable of operating normally on the small intestine dataset, performing real-time registration at 250 frames per second. Additionally, it has significantly shorter fine-tuning and retraining times compared to other learning-based methods, demonstrating superior efficiency and adaptability.

\section{conclusion}
In this paper, we introduced GERA, a method for real-time point cloud processing utilizing an offline geometric information constructor. We conducted experiments using MMD and a challenging small bowel dataset to validate the robustness and superiority of our approach. However, our experiments were primarily focused on the registration of individual organ data and did not consider the registration of composite objects in scenarios like scene flow. Our future research direction is to apply GERA's offline geometric encoder to the scene flow problem, aiming to achieve high-precision scene flow estimation with no time delay.

\bibliographystyle{IEEEtran}
\bibliography{mybib}


\end{document}